\documentclass[runningheads]{llncs}
\usepackage[T1]{fontenc}
\usepackage{latexsym}
\usepackage{amssymb}
\usepackage[nonamelimits]{amsmath}
\usepackage{graphicx}
\usepackage{xcolor}
\usepackage[pdftex,
              pdftitle={Invariance assumptions for class distribution estimation},
               pdfsubject={Machine learning},
               pdfkeywords={},
               pdfauthor={Dirk Tasche},
               pdfstartview=FitR,
               breaklinks=true,
               colorlinks=true,
               citecolor=teal,
               linkcolor=magenta,
               urlcolor=olive]{hyperref}
\usepackage[authoryear, round, sectionbib]{natbib}

\renewcommand{\bibname}{References}

\begin{document}
\title{Invariance assumptions for class distribution estimation}
%
%
\author{Dirk Tasche\orcidID{0000-0002-2750-2970}}
\authorrunning{D.~Tasche}
%
\institute{Independent Scholar, \email{dirk.tasche@gmx.net}\\
First version: May 25, 2023\\
This version: August 27, 2023}
\maketitle              
\begin{abstract}
We study the problem of class distribution estimation under dataset shift. 
On the training dataset, both features and class labels are observed while on the test dataset only 
the features can be observed. The task then is the estimation of the distribution of the class labels, i.e.\ the
estimation of the class prior probabilities, in the test dataset. Assumptions of invariance between
the training joint distribution of features and labels and the test distribution can considerably facilitate this task.
We discuss the assumptions of covariate shift, factorizable joint shift, and sparse joint shift and
their implications for class distribution estimation.

\end{abstract}

\section{Introduction}

We consider class distribution estimation against the backdrop of dataset shift 
(also called distribution shift) between training 
and test dataset. On the training dataset, both features and class labels are observed while on the test dataset only 
the features can be observed. In this context, important tasks of interest are the prediction of the
labels (classification) and the estimation of the label distribution (class distribution
estimation) in the test dataset. In the literature, class distribution estimation is also referred to as class prior estimation,
class prevalence estimation, quantification, and with a number of other terms.

Referring to \citet{forman2005counting},  \citet[][Preface]{esuli2023learning} 
made the following case for class distribution estimation as a research topic of its own: 
``In a number of applications involving classification, 
the final goal is not determining
which class (or classes) individual unlabelled instances belong to, but estimating
the prevalence (or `relative frequency', or `prior probability') of each class in the
unlabelled data.'' 

Class distribution estimation for the target (test) dataset when its distribution is allowed to differ from 
the distribution of the training (source) dataset, in general, is an ill-posed problem,
because joint target (test) distributions of features and labels whose marginal feature distributions 
perfectly match the observed target feature distribution cannot be distinguished. 
Constraints are needed on the range of joint target distributions taken into account for the estimation
exercise in order to make the problem well-posed.
The consideration of causality is a popular approach for specifying such constraints. Typically,
this approach leads to making a decision either for prior probability shift (label shift) or for covariate shift
as the model for the joint target distribution \citep{fawcett2005response}. 

Other approaches to the problem include
\vspace{-1ex}
\begin{itemize}
 \item Assumptions on the evolution of parts of the joint distribution of labels and features between
 training and test times \citep[e.g.][]{Zhang:2013:TargetShift,krempl2019temporal}.
\item Implicit assumptions, for instance by the choice of the distance function for measuring the difference of
the source and the target feature distributions \citep[e.g.][]{hofer2015adapting,kirchmeyer2021mapping}.
\end{itemize} 
In this paper, we revisit three approaches to class distribution estimation and, more generally, to
modelling dataset shift under invariance assumptions between the joint source and target distributions: 
Covariate shift \citep{shimodaira2000improving},
factorizable joint shift \citep[FJS,][]{he2022domain}, and sparse joint shift \citep[SJS,][]{chen&zaharia&Zou:SJS}.

The contribution of this paper to the literature is twofold. On the
one hand, two new approaches to class distribution estimation under covariate shift are presented. These approaches may
prove useful for cross-checking estimates obtained by application of the popular `probabilistic classify and count'
approach. On the other hand, some results on FJS and SJS which were presented in \citet{tasche2022factorizable}
and \citet{tasche2023sparse} in uncommon notation are revisited in a notation more familiar to the machine learning
community.

Class distribution estimation under prior probability shift has been receiving a lot of attention 
by the research community for at least the last sixty years, beginning with \citet{buck1966comparison}
if not earlier. 
For this reason, in this paper we do not dive into any detail of prior probability shift. Regarding this topic, we
refer to the recent overviews by \citet{Gonzalez:2017:RQL:3145473.3117807} and \citet{esuli2023learning} 
of the literature on class distribution estimation under prior probability shift and the references therein.

This paper is organised as follows:
\vspace{-1ex}
\begin{itemize}
\item Section~\ref{se:Setting} `Notation and general assumptions' sets the scene in technical terms
for the remainder of the paper.
\item Section~\ref{se:definitions} `Types of dataset shift with invariance assumptions' 
provides the formal definitions of the four most important types of distribution
shift considered in more or less detail in the following: 
Prior probability shift, covariate shift, factorizable joint shift (FJS), and sparse joint shift (SJS).
\item Section~\ref{se:CovSH} `Covariate shift' looks at class distribution estimation under
covariate shift, based on previous work by \citet{card2018importance} and \citet{tasche2022class}. Eq.~\eqref{eq:wEst}
and Proposition~\ref{pr:noSufficiency} are new results.
\item Section~\ref{se:FactSh} `Factorizable joint shift (FJS)' revisits the notion of distribution shift
proposed by \citet{he2022domain}. FJS is found to be unsuitable for class distribution estimation due 
to lack of identifiability unless additional constraints are applied.
\item Section~\ref{se:SpSh} `Sparse joint shift (SJS)' summarises findings of \citet{chen&zaharia&Zou:SJS}
and \citet{tasche2023sparse}. Proposition~\ref{pr:condConfusion} on the `conditional confusion matrix approach'
presents a new interpretation of a result of \citet{tasche2023sparse}. SJS is shown to be
a generalisation of prior probability shift and found to be a suitable assumption for designing
class distribution estimators.
\item The paper concludes with a brief assessment of the findings in Section~\ref{se:Concl}.
\end{itemize}

\section{Notation and general assumptions}
\label{se:Setting}

We adopt notation and assumptions similar to the setting used in \citet{Scott2019}:

There are a feature space $\mathcal{X}$ (not necessarily with $\mathcal{X} \subset \mathbb{R}^d$ 
for any fixed $d$) and a
label space $\mathcal{Y} = \{1, \ldots, \ell\}$ for some integer $\ell \ge 2$. This is the common machine learning
setting for  multinomial classification and class distribution estimation.

As in \citet[][Section~1.2]{Scott2019}, ``$\ldots$ there are two distributions, $P$ and $Q$, 
referred to as the \emph{source and target distribution}s. We consider the semi-supervised setting 
where the learner observes $(X_1, Y_1), \ldots, (X_m, Y_m) \sim P$ and $X_{m+1}, \ldots, X_{m+n} \sim Q_X$ $\ldots$''.

$P$, $Q$ are probability distributions on $\mathcal{X} \times \mathcal{Y}$.
$P$ is also called \emph{training distribution}, $Q$ \emph{test distribution}.
$X$ is a generic random variable which shows the features of an object (or instance), $Y$ is a generic random
variable showing the class label of an object. $Q_X$ stands for the marginal distribution of the features under
the target distribution.

We suppose for the purpose of this paper that the sample sizes $m$ of the training sample
and $n$ of the test sample are sufficiently large if not infinite such that $P$ and $Q_X$ can be 
perfectly inferred and assumed to be known.

Class distribution estimation then may be phrased as the problem of how to find 
the marginal distribution $Q_Y$ of the labels (i.e.\ the class distribution)
under the target distribution, i.e.\ the prior probabilities $Q[Y=1]$, $\ldots$, $Q[Y=\ell]$.
\vspace{-2ex}

\subsubsection{Densities.}
In the following, we assume that the joint target distribution $Q$ of features and labels $(X,Y)$ 
is absolutely continuous
\citep[see][Definition~7.30]{klenke2013probability} with respect to the joint source distribution $P$ of $(X,Y)$. 
We also suppose that $p = p(x, y)$ is a joint density of $(X,Y)$ under $P$ and 
$q = q(x, y)$ is a joint density of $(X,Y)$ under $Q$, with respect to some third measure.
Absolute continuity of $Q$ with respect to $P$ is implied in particular if
the support of $Q$ is a subset of the support of $P$, i.e.\ if it holds that
\begin{equation}\label{eq:abscont}
q(x, y) > 0 \quad \Rightarrow \quad p(x,y) > 0.
\end{equation}
For the sake of simplying the notation, for the remainder of the paper we assume that \eqref{eq:abscont} is true.

\begin{subequations}
Under the assumption that \eqref{eq:abscont} holds, define the general \emph{importance weight} function 
$w(x,y)$ for $x \in \mathcal{X}$ and $y \in \mathcal{Y}$ by
\begin{equation}\label{eq:wxy}
w(x,y) = \begin{cases}
\frac{q(x,y)}{p(x,y)}, & \text{for}\ p(x,y) >0, \\
0, & \text{for}\ p(x,y) = 0.
\end{cases}
\end{equation}
Function $w$ reflects the change caused by transitioning from source $P$ to target $Q$. It can
also be interpreted as the density of $Q$ with respect to $P$ on $\mathcal{X} \times \mathcal{Y}$.

Besides the full densities $p$ and $q$ also the marginal densities $p_X$, $q_X$ of the feature variable $X$ 
are of interest:
\begin{equation*}
p_X(x)  = \sum_{y=1}^\ell p(x,y), \quad
q_X(x)  = \sum_{y=1}^\ell q(x,y).
\end{equation*}
The feature densities $p_X$, $q_X$ give rise to the \emph{feature importance weight} function 
$w_X(x)$ for $x \in \mathcal{X}$ which is defined by
\begin{equation}\label{eq:wX}
w_X(x) = \begin{cases}
\frac{q_X(x)}{p_X(x)}, & \text{for}\ p_X(x) >0, \\
0, & \text{for}\ p_X(x) = 0.
\end{cases}
\end{equation}
\end{subequations}
\vspace{-2ex}

\subsubsection{Posterior probabilities.}
We denote the \emph{posterior probability} (conditional probability) of class $y \in \mathcal{Y}$ 
given the feature variable $x$ under the source distribution $P$ by $P[Y=y\,|\,X=x]$. This is a single number.
$P[Y=y\,|\,X]$ stands for the  random variable created by sampling $x$ from the feature distribution $P_X$
and evaluating $P[Y=y\,|\,X=x]$ at $x$.\\
$Q[Y=y\,|\,X=x]$ and $Q[Y=y\,|\,X]$ respectively denote the corresponding posterior probabilities under
the target distribution $Q$.

Recall also the definition of the \emph{class-conditional feature distributions} $P_{Y=y}$ and $Q_{Y=y}$ 
under the source distribution $P$ and target distribution $Q$ respectively by
\begin{equation}\label{eq:classCond}
\begin{split}
P_{Y=y}[X\in M] & = P[X\in M\,|\,Y=y] = \frac{P[X\in M, Y=y]}{P[Y=y]},\\
Q_{Y=y}[X\in M] & = Q[X\in M\,|\,Y=y] = \frac{Q[X\in M, Y=y]}{P[Y=y]},
\end{split}
\end{equation}
for $M \subset \mathcal{X}$.
\vspace{-2ex}

\subsubsection{Further notation.} In the following, we denote by $\mathbf{C} = (C_1, \ldots, C_\ell)$ 
hard \emph{multinomial classifiers} in the sense that
\begin{equation}\label{eq:classifier}
\begin{split}
 & C_i \subset \mathcal{X} \ \text{for all}\ i = 1, \ldots, \ell,  \\
& C_1, \ldots, C_\ell\ \text{is a disjoint decomposition of}\ \mathcal{X},\ \text{and}  \\
& Y = y\ \text{is predicted when}\ X \in C_y\ \text{is observed.}
\end{split}
\end{equation}
The \emph{indicator function} $\mathbf{1}_S$ of a set $S$ is defined as
$\mathbf{1}_S(s)=1$ for $s\in S$ and $\mathbf{1}_S(s)=0$ for $s\notin S$.

\section{Types of dataset shift with invariance assumptions}
\label{se:definitions}

This section formally introduces the types of dataset shift to be discussed in the remainder of the
paper.

The dataset shift type denoted here by prior probability shift is also called label shift, target shift, global drift, 
or named in other ways in the literature. Under this type of shift, the class-conditional feature
distributions are invariant between source and target distribution. Its definition is given here
mainly as a point of reference.
\begin{definition}[Prior Probability Shift]\label{de:prior}
For each $y \in \mathcal{Y}$, the class-conditional feature distributions
$P_{Y=y}[X \in M]$ and $Q_{Y=y}[X \in M]$ for measurable $M \subset \mathcal{X}$ as defined 
by \eqref{eq:classCond} are equal, i.e.\ it holds that
\begin{equation*}
P_{Y=y}[X \in M] = Q_{Y=y}[X \in M], \quad \text{for}\ y \in \mathcal{Y},\ M \subset \mathcal{X}.
\end{equation*}
\end{definition}
The notion of covariate shift was introduced by \citet{shimodaira2000improving}. 
It is based on the possibly most popular invariance assumption for the relationship between
source distribution and target distribution: The posterior class probabilities (sometimes called
the `concept') remain unchanged.
We quote mutandis mutatis the definition of covariate shift from \citet{kpotufe2021covariate}.
\begin{definition}[Covariate Shift]\label{de:covSh}
For each $y \in \mathcal{Y}$, there exists a measurable function $\eta_y: \mathcal{X} \to [0,1]$, 
called \emph{posterior class probability}, such that 
\begin{equation}\label{eq:covSh}
P[Y=y\,|\,X=x] = \eta_y(x) = Q[Y=y\,|\,X=x],
\end{equation}
almost surely for all $x$ under $P_X$ and under $Q_X$.
\end{definition}
Class distribution estimation in the presence of covariate shift is discussed below in Section~\ref{se:CovSH}.

Against the backdrop that, under the assumptions of this paper, it is impossible to distinguish prior probability shift
and covariate shift solely on the basis of data, the following notion of factorizable joint shift (FJS) as 
proposed by \citet{he2022domain} is very appealing at first glance. For
it includes both prior probability shift and covariate shift as special cases and, thus, may be
interpreted as interpolating between these two poles of dataset shift. 

\begin{subequations}
\begin{definition}[Factorizable joint shift (FJS)] \label{de:FJS}
There exist non-negative functions $u$ on $\mathcal{X}$ and $v$ on $\mathcal{Y}$ such that
for the importance weight function $w$ as defined in \eqref{eq:wxy}, it holds that
\begin{equation}\label{eq:uv}
w(x,y) = u(x)\,v(y),
\end{equation}
 almost surely for all $(x,y) \in \mathcal{X}\times \mathcal{Y}$ under $P$.
\end{definition}
Observe that the functions $u$ and $v$ of Definition~\ref{de:FJS} are not uniquely determined 
because for any $c>0$ the functions $u_c = c\,u$ and $v_c = v / c$ also satisfy \eqref{eq:uv}:
\begin{equation}\label{eq:uvc}
w(x,y) = u_c(x)\,v_c(y).
\end{equation}
\end{subequations}
No invariance property between the source and target distributions is obvious from Definition~\ref{de:FJS}.
Such a property, nonetheless, is implied by Theorem~\ref{th:factorized} below in Section~\ref{se:FactSh}
which is devoted to a discussion of FJS.

\citet{chen&zaharia&Zou:SJS} proposed ``a new distribution shift model, Sparse Joint Shift (SJS),
which considers the joint shift of both labels and a few features. This unifies and generalizes 
existing shift models including label shift and sparse covariate shift\footnote{%
See Definition~\ref{de:SCS} below for a definition of sparse covariate shift.}, 
where only marginal feature or label distribution shifts are considered.''

\begin{definition}[Sparse Joint Shift (SJS)]\label{de:SJS}
Let $T:\mathcal{X}\to \mathcal{T}$ be a measurable transformation of the feature values $x$.
The source distribution $P$ and the target distribution $Q$ are related through 
$T$-SJS if it holds for all $y \in \mathcal{Y}$ and $M \subset \mathcal{X}$ that
\begin{equation}\label{eq:SJS}
P_{Y=y}[X \in M\,|\,T(X)=t] = Q_{Y=y}[X \in M\,|\,T(X)=t]
\end{equation}
for all $t\in \mathcal{T}$ almost surely under $P_{T(X)}$ and $Q_{T(X)}$.
\end{definition}
Under SJS, the doubly conditioned (by class and by a transformation of the features) feature distributions
are invariant between source distribution and target distribution. Note that $T(X)$ in general 
creates a `sparse' or `thinned out' version of the features. \citet[][Section~3.1]{chen&zaharia&Zou:SJS} called 
this type of shift `sparse' because ``the sparsity is necessary for the shift to be identifiable''.\\
Choosing $T$ in Definition~\ref{de:SJS} as $T(x) = c$ for all $x \in \mathcal{X}$, where $c$ is some fixed
value, shows that prior probability shift in the sense of Definition~\ref{de:prior} is a special case of
SJS. In certain limited circumstances, covariate shift implies SJS and vice versa, as is discussed
below in Section~\ref{se:SpSh}. In general, however, covariate shift is not a special case of SJS.\\
If $P$ and $Q$ are related through an `exponential tilt model' as defined in Section~3 of 
\citet{maity2023understanding} then $P$ and $Q$ are also related through SJS.

\section{Covariate shift}
\label{se:CovSH}

This section gives a brief overview of class distribution estimation under covariate shift. 
The topic appears to not have received much attention in the literature,
with the exceptions of \citet{card2018importance} and \citet{tasche2022class}. 
\vspace{-2ex}

\subsubsection{Class prior estimators.}
If $\mathbf{C}=(C_1, \ldots, C_\ell)$ is a multinomial classifier as defined by \eqref{eq:classifier}, 
\emph{classify \& count} \citep{forman2005counting} might be the most obvious
class prior estimator $\widetilde{Q}_n[Y=y]$, $y = 1, \ldots, \ell$, under any type of 
dataset shift:
\begin{equation*}
\widetilde{Q}_n[Y=y] = \frac{1}{n} \sum_{i=1}^n \mathbf{1}_{C_y}(x_i),
\end{equation*}
where $x_1, \ldots, x_n$ is a test sample of feature values, assumed to have been generated with
the target feature distribution $Q_X$. If $x_1, \ldots, x_n$ is an i.i.d.\ sample from $Q_X$, it follows
that $\widetilde{Q}_n[Y=y] \to Q[X \in C_y]$ for $n \to \infty$. However, given that
$Q_X$ may be any distribution on $\mathcal{X}$, under covariate shift there is no reason why
$Q[X \in C_y]$ should equal $Q[Y=y]$ unless $\mathbf{C}$ is a perfect classifier under the target 
distribution $Q$ -- which is an unrealistic assumption. 

As noted by \citet{card2018importance}, valid estimates $\widehat{Q}_n[Y=y]$ of 
the target prior probabilities
$Q[Y=y]$, $y = 1, \ldots, \ell$, under covariate shift can be obtained by 
taking recourse to the law of total probability. 
The law of total probability implies
\begin{subequations}
\begin{equation}\label{eq:totallaw}
Q[Y=y] = E_Q\bigl[P[Y=y\,|\,X]\bigr] = \int_\mathcal{X} P[Y=y\,|\,X=x]\,Q_X(dx).
\end{equation}
This gives the estimator
\begin{equation}\label{eq:emp}
\widehat{Q}_n[Y=y] = \frac{1}{n} \sum_{i=1}^n \widehat{P}[Y=y\,|\,X=x_i],
\end{equation}
where $x_1, \ldots, x_n$ is a test sample of feature values, as described above, 
and $\widehat{P}[Y=y\,|\,X=x]$ denotes an estimate of the
posterior probability $P[Y=y\,|\,X=x]$ under the source distribution $P$, evaluated at the feature value $x$.
Estimator \eqref{eq:emp} was called \emph{probabilistic classify and count (PCC)} by \citet{card2018importance}
and \emph{probability estimation \& average (P\&{}A)} by \citet{bella2010quantification}.
\end{subequations}

With the feature importance weight function $w_X$ defined by \eqref{eq:wX}, under covariate shift it holds true that 
\begin{subequations}
\begin{equation}
Q[Y=y]= E_P[w_X(X)\,\mathbf{1}_{\{y\}}(Y)], \quad y \in \mathcal{Y}.
\end{equation}
Hence, once the importance weight function $w_X$ has been estimated from a sample of features
generated under $P$ and another sample of features generated under $Q$, the class prior probabilities
$Q[Y=y]$ can be estimated by means of the estimator 
\begin{equation}\label{eq:wEst}
\bar{Q}_m[Y=y] = \frac{1}{m} \sum_{i=1}^m w_X(x_i)\,\mathbf{1}_{\{y\}}(y_i),
\end{equation}
where $(x_1, y_1), \ldots, (x_m, y_m)$ is an i.i.d.\ sample of $(X,Y)$ under the source distribution $P$.
A variety of methods is available for estimating $w_X$, see e.g.\ \citet{sugiyama2012density} or
\citet{bickel2009discriminative}. \citet{card2018importance} might have deployed estimator 
\eqref{eq:wEst}, calling it \emph{reweighting} estimator. They did not, however, provide an explicit formula for it. 
A potential application of \eqref{eq:wEst} would be to make use of it for cross-checking primary estimates
of the target prior probabilities resulting from an application of \eqref{eq:emp}.
\end{subequations}
\vspace{-2ex}

\subsubsection{Dimension reduction.} $X$ may be a high dimensional random vector such that precisely 
estimating $x \mapsto P[Y=y\,|\,X=x]$ is
difficult, and also the computation of the high-dimensional integral on the right-hand side of \eqref{eq:totallaw}
is a hard task. Hence, is it possible to reduce the dimension of $X$ by applying a transformation $T$ such that
$T(X)$ has a lower dimension than $X$ but some version of \eqref{eq:totallaw}, e.g.\ like 
\eqref{eq:totallawZ}, still holds true:
\begin{subequations}
\begin{equation}\label{eq:totallawZ}
Q[Y=y] \stackrel{\text{\large{}?}}{=} E_Q\bigl[P[Y=y\,|\,T(X)]\bigr] = 
\int_\mathcal{T} P[Y=y\,|\,T(X)=t]\,Q_{T(X)}(dt),
\end{equation}
supposing that the transformation $T$ takes its values in $\mathcal{T}$.

\citet[][Theorem~1]{tasche2022class} showed that 
\begin{equation}\label{eq:TX}
P[Y=y\,|\,T(X)] = Q[Y=y\,|\,T(X)]
\end{equation}
\end{subequations}
is true under covariate shift with the same transformation $T(X)$ 
for all target distributions $Q$ which are absolutely continuous with respect to the fixed source distribution $P$ 
if and only if 
\begin{equation}\label{eq:sufficiency}
P[Y=y\,|\,T(X)=T(x)] = P[Y=y\,|\,X=x],
\end{equation}
almost surely for all $x$ under $P_X$. \eqref{eq:sufficiency} means that $T(X)$ is \emph{sufficient} 
for $X$ with respect
to $Y=y$ \citep[see][Section~3]{Tasche2022}. In general, 
requesting sufficiency for $T(X)$ excludes simple approaches to dimension reduction for $X$.
Hence, most of the time there is no guarantee that \eqref{eq:TX} and consequently also \eqref{eq:totallawZ} are
applicable.

Although \eqref{eq:TX} is not true in general without an assumption of sufficiency, 
thanks to the generalised Bayes' theorem \citep[][Theorem~10.8]{Klebaner} 
covariate shift can still be shown to imply the following variation of \eqref{eq:covSh}
for a fixed target distribution $Q$:
\begin{proposition}\label{pr:noSufficiency}
Suppose that $Q$ is absolutely continuous with respect to $P$ and $Q$ and $P$ are related through
covariate shift in the sense of Definition~\ref{de:covSh}. Then it follows for any measurable transformation
$T:\mathcal{X}\to\mathcal{T}$ and all $y\in \mathcal{Y}$ that
\begin{equation*}
Q[Y=y\,|\,T(X)=t] = \frac{E_P\bigl[w_X(X)\,\mathbf{1}_{\{y\}}(Y)\,|\,T(X)=t]\bigr]}
    {E_P[w_X(X)\,|\,T(X)=t]},
\end{equation*}
for all $t\in \mathcal{T}$ almost surely under $P_{T(X)}$, where $w_X$ is defined as in \eqref{eq:wX}.
\end{proposition}

As a consequence of Proposition~\ref{pr:noSufficiency}, 
\eqref{eq:TX} holds true for fixed $Q$ if and only if
\begin{equation}\label{eq:condind}
E_P\bigl[w_X(X)\,\mathbf{1}_{\{y\}}(Y)\,|\,T(X)] = E_P[w_X(X)\,|\,T(X)]\, P[Y=y\,|\,T(X)],
\end{equation}
i.e.\ if $w_X(X)$ and $\{Y=y\}$ are independent conditional on $T(X)$ under $P$. Such conditional independence,
in particular, follows if $T(X)$ is sufficient for $X$ with respect to $\{Y=y\}$.
Accordingly, in principle
it is possible to check by means of verification of \eqref{eq:condind}
whether or not \eqref{eq:totallawZ} can be applied.
This involves the estimation of $w_X$ which, at first glance, might not be much easier or even harder than 
estimating $P[Y=y\,|\,X]$.

See, however, \citet[][Section~3]{stojanov2019low} for a method to identify a transformation $T$ such
that $T(X)$ is approximately sufficient for $X$ with respect to all $\{Y=y\}$, $y \in \mathcal{Y}$.
By \eqref{eq:condind}, then \eqref{eq:TX} holds for the target distribution $Q$ in question such that 
\eqref{eq:totallawZ} is applicable.

\section{Factorizable joint shift (FJS)}
\label{se:FactSh}

\citet{he2022domain} characterised FJS by claiming that 
``the biases coming from the data and the label are statistically independent'', without specifying 
any detail of the claim in technical terms. \citet{tasche2022factorizable} suggested that
FJS might be interpreted as a structural property similar to the `separation of variables' 
which plays an important role for finding closed-form solutions to differential equations.

As noted by \citet{he2022domain}, covariate shift is a special case of FJS because of
\begin{subequations}
\begin{equation}
w(x,y) = w_X(x)
\end{equation}
for $w_X$ defined by \eqref{eq:wX}, and prior probability shift is a special case of FJS because of
\begin{equation}
w(x,y) =  \frac{Q[Y=y]}{P[Y=y]}.
\end{equation}
\end{subequations}
\vspace{-2ex}

\subsubsection{Characterising FJS.}
\citet{he2022domain} also noted that
FJS is not fully identifiable in the unsupervised setting of this paper, i.e.\ if
no labels are observed in the target dataset. In the remainder of this section, we summarise the analysis 
of FJS performed by \citet{tasche2022factorizable} and clarify the additional 
assumptions needed to achieve identifiability for FJS. 

The following theorem implies, among other things,  an invariance property between
source distribution $P$ and target distribution $Q$ thanks to FJS (see Eq.~\eqref{eq:ratio} below).

\begin{theorem}\label{th:factorized}
Suppose that the source distribution $P$ and the target distribution $Q$ are 
related by FJS in the sense of Definition~\ref{de:FJS}.
Denote by $w_X$ the feature importance weight function defined by \eqref{eq:wX} and
let $q_i = Q[Y=i]$ and $p_i = P[Y=i]$, $i = 1, \ldots, \ell$.\\
Then, up to a constant factor $c$ as in \eqref{eq:uvc}, it follows that
\begin{subequations}
\begin{align}\label{eq:v}
v(y) & = \sum_{i=1}^{\ell-1} \varrho_i\,\frac{q_i}{p_i}\,\mathbf{1}_{\{i\}}(y) +
\frac{q_\ell}{p_\ell}\,\mathbf{1}_{\{\ell\}}(y)\quad \text{and}\\
u(x) & = \frac{w_X(x)}{\sum_{i=1}^{\ell-1} \varrho_i\,\frac{q_i}{p_i}\,P[Y=i\,|\,X=x] +
\frac{q_\ell}{p_\ell}\,P[Y=\ell\,|\,X=x]},\label{eq:u}
\end{align}
where the constants $\varrho_1, \ldots, \varrho_{\ell-1}$ are positive and finite and 
satisfy the following equation system (with $j = 1, \ldots, \ell-1$):
\begin{equation}\label{eq:system}
p_j  =  \varrho_j\,E_P\left[\frac{w_X(X)\,P[Y=j\,|\,X]}
{\sum_{i=1}^{\ell-1} \varrho_i\,\frac{q_i}{p_i}\,P[Y=i\,|\,X] +
\frac{q_\ell}{p_\ell}\,P[Y=\ell\,|\,X]}\right].
\end{equation}
\end{subequations}
Conversely, suppose that for the source distribution $P$ a function $w_X:\mathcal{X}\to[0,\infty)$
 with $E_P\bigl[w_X(X)\bigr] =1$ 
and $(q_i)_{i=1, \ldots, \ell} \in (0,1)^\ell$ with $\sum_{i=1}^\ell q_i =1$ are given. 
Assume also that $\varrho_1 >0$, $\ldots$, $\varrho_{\ell-1} > 0$ are solutions of 
the equation system \eqref{eq:system} 
and $u$ and $v$ are defined by \eqref{eq:u} and \eqref{eq:v}, respectively. 
Then $w(x,y) = u(x)\,v(y)$ has the property that $w(x,y)\,p(x,y)$ is the 
density of a probability measure $Q$ on $\mathcal{X}\times \mathcal{Y}$ such that $w_X(x)\,p_X(x)$ is the
marginal density of the feature variable $X$  under $Q$ 
and $Q[Y=i] = q_i$ holds for $i = 1, \ldots, \ell$.
\end{theorem}
See \citet[][Theorem~2]{tasche2022factorizable} for a proof of Theorem~\ref{th:factorized}.
The theorem characterises FJS through equations \eqref{eq:u}, \eqref{eq:v} and \eqref{eq:system} but
does not provide any information regarding the existence or uniqueness of solutions to \eqref{eq:system}.
A result on existence and uniqueness of the solutions to \eqref{eq:system} was proven for the binary
case $\ell = 2$ by \citet[][Proposition~2]{tasche2022factorizable}.

It can be shown \citep[][Corollary~4]{tasche2022factorizable} that Theorem~\ref{th:factorized} implies
the following version of the correction formula for class posterior probabilities 
of \citet[][Eq.~(2.4)]{saerens2002adjusting} and \citet[][Theorem~2]{Elkan01} under FJS.
\begin{corollary}\label{co:factorized}
Suppose that the source distribution $P$ and the target distribution
$Q$ are related through FJS in the sense of Definition~\ref{de:FJS}. Then the
target posterior probabilities $Q[Y=j\,|\,X=x]$, $j = 1, \ldots, \ell$, can be represented
almost surely for all $x$ under $Q_X$ as functions of
the source posterior probabilities $P[Y=j\,|\,X=x]$, $j = 1, \ldots, \ell$, in
the following way:
\begin{equation*}
\begin{split}
Q[Y=j\,|\,X=x] & = \frac{\varrho_j\,\frac{Q[Y=j]}{P[Y=j]} P[Y=j\,|\,X=x]}
{\sum_{i=1}^{\ell-1} \varrho_i\,\frac{Q[Y=i]}{P[Y=i]} P[Y=i\,|\,X=x] +
\frac{Q[Y=\ell]}{P[Y=\ell]} P[Y=\ell\,|\,X=x]},\\
& \qquad j = 1, \ldots, \ell-1,\\
Q[Y=\ell\,|\,X=x] & = \frac{\frac{Q[Y=\ell]}{P[Y=\ell]} P[Y=\ell\,|\,X=x]}
{\sum_{i=1}^{\ell-1} \varrho_i\,\frac{Q[Y=i]}{P[Y=i]} P[Y=i\,|\,X=x] +
\frac{Q[Y=\ell]}{P[Y=\ell]} P[Y=\ell\,|\,X=x]},
\end{split}
\end{equation*}
where the positive constants $\varrho_1, \ldots, \varrho_{\ell-1}$ satisfy the equation 
system \eqref{eq:system}.
\end{corollary}

Corollary~\ref{co:factorized} in turn implies that under FJS the following invariance property holds true:
\begin{equation}\label{eq:ratio}
\frac{Q[Y=j\,|\,X]}{Q[Y=\ell\,|\,X]} \frac{Q[Y=\ell]}{Q[Y=j]}  = 
\varrho_j\, \frac{P[Y=j\,|\,X]}{P[Y=\ell\,|\,X]} \frac{P[Y=\ell]}{P[Y=j]},
\quad j = 1, \dots, \ell-1,
\end{equation}
where the constants $\varrho_j$ satisfy the equation system \eqref{eq:system}.
Eq.~\eqref{eq:ratio} may be interpreted as stating that
under factorizable joint shift the ratios of the class-conditional feature densities are invariant 
between source and target distributions up
to a constant factor \citep[see][Remark~1]{tasche2022factorizable}.
\vspace{-2ex}

\subsubsection{Class distribution estimation under FJS.} 
Theorem~\ref{th:factorized} suggests two obvious ways to 
learn the characteristics of factorizable joint shift: 
\begin{itemize}
\item[a)] If the target prior class probabilities $Q[Y=i]=q_i$ are known (for instance from external sources)
solve \eqref{eq:system} for the constants $\varrho_i$.
\item[b)] If the target prior class probabilities $Q[Y=i]=q_i$ are unknown (as would be the case for
the problem of class distribution estimation), fix values for the
constants $\varrho_i$ and solve \eqref{eq:system} for the $q_i$. Letting $\varrho_i =1$ for
all $i$ is a natural choice that converts \eqref{eq:system} into the system of maximum likelihood
equations for the $q_i$ under the prior probability shift assumption. 
\end{itemize}
See Section~4.2.4 of \citet{tasche2013art} for an example of approach~a) from the area of credit risk.
Whenever for a given marginal target feature distribution $Q_X$ there is more than one set of potential
target class prior probabilities $q_y$, $y = 1, \ldots, \ell$, such that \eqref{eq:system} can be 
solved for the $\varrho_i$, then a case of unidentifiability of the joint target distribution $Q$ 
under FJS is incurred. This always holds for the binary case $\ell = 2$ because for any given combination
of joint source distribution $P$, target feature distribution $Q_X$ and target prior 
probability $q_1 = Q[Y=1]$, a constant $\varrho_1$ can be found such
that $P$ and $Q$ are related through FJS \citep[][Proposition~2]{tasche2022factorizable}.

Regarding the interpretation of \eqref{eq:system} in approach~b) as maximum likelihood equations, see
\citet{duPlessis2014110}. This interpretation, in particular, implies that
an EM (expectation maximisation) algorithm can be deployed for solving the equation 
system \citep{saerens2002adjusting} in the case $1 = \varrho_1 = \ldots = \varrho_{\ell-1}$.
 
\section{Sparse joint shift (SJS)}
\label{se:SpSh}

Definition~\ref{de:SJS} of SJS slightly generalises Definition~1 of \citet{chen&zaharia&Zou:SJS} as
can be seen by choosing $T$ as extractor of a subset of the components of the feature vector. The equivalence
of this special case of Definition~\ref{de:SJS} and the definition of \citet{chen&zaharia&Zou:SJS} then
follows from Proposition~3.8 of \citet{tasche2023sparse}. 

Observe that by the generalised Bayes' theorem \citep[][Theorem~10.8]{Klebaner}, 
\eqref{eq:SJS} can equivalently be stated as
\begin{equation}\label{eq:SJS2}
\frac{P[X \in M, Y=y\,|\,T(X)=t]}{P[Y=y\,|\,T(X)=t]} = \frac{Q[X \in M, Y=y\,|\,T(X)=t]}{Q[Y=y\,|\,T(X)=t]}.
\end{equation}

The following properties of SJS were first noted by \citet{tasche2023sparse}.
\begin{proposition}[Properties of SJS]\label{pr:SJS}
Suppose that the source distribution $P$ and the target distribution $Q$ are related through
$T$-SJS in the sense of Definition~\ref{de:SJS}. Then the following two statements
hold true:
\begin{itemize}
\item[(i)] If $T':\mathcal{X} \to \mathcal{T}'$ and $S:\mathcal{T}'\to \mathcal{T}$ are measurable
transformations such that for all $x\in \mathcal{X}$ it holds that $T(x) = (S\circ T')(x) =
S\bigl(T'(x)\bigr)$, then $P$ and $Q$ are also related through $T'$-SJS.
\item[(ii)] For all $i \in \mathcal{Y}$, it holds that
\begin{equation*}
Q[Y=i\,|\,X=x] = \frac{\frac{Q[Y=i\,|\,T(X)=T(x)]}{P[Y=i\,|\,T(X)=T(x)]}\,P[Y=i\,|\,X=x]}
{\sum_{j=1}^\ell \frac{Q[Y=j\,|\,T(X)=T(x)]}{P[Y=j\,|\,T(X)=T(x)]}\,P[Y=j\,|\,X=x]},
\end{equation*}
for all $x \in \mathcal{X}$ almost surely under $Q_X$.
\end{itemize}
\end{proposition}

See \citet[][Corollary~4.3]{tasche2023sparse} for a proof of Proposition~\ref{pr:SJS}~(i) and
\citet[][Proposition~4.5]{tasche2023sparse} for a proof of Proposition~\ref{pr:SJS}~(ii).
By Proposition~\ref{pr:SJS}~(i), prior probability shift implies $T$-SJS for any transformation 
$T:\mathcal{X}\to\mathcal{T}$.
Proposition~\ref{pr:SJS}~(ii) is another generalisation of the posterior correction formula of 
\citet[][Eq.~(2.4)]{saerens2002adjusting} and \citet[][Theorem~2]{Elkan01}, this time under the
assumption of SJS.

The next result rephrases the identifiability result of \cite[][Theorem~1]{chen&zaharia&Zou:SJS}
in terms of conditional expectations instead of joint densities.
\begin{theorem}[Identifiability under SJS]\label{th:SJS}
Suppose that there are distributions $P$, $Q$ and $Q'$ on $\mathcal{X}\times\mathcal{Y}$  as well
as transformations $T:\mathcal{X}\to\mathcal{T}$ and $T':\mathcal{X}\to\mathcal{T}'$ such that
$P$ and $Q$ are related through $T$-SJS and $P$ and $Q'$ are related through $T'$-SJS.
For given measurable functions $f_i:\mathcal{X}\to[0,\infty)$, $i = 1, \ldots, \ell$, define
the random matrix $R(X) = \bigl(R_{ij}(X)\bigr)_{i,j\in\{1, \ldots, \ell\}}$ by
\begin{equation*}
R_{ij}(X) = \frac{E_P\bigl[f_i(X)\,\mathbf{1}_{\{j\}}(Y)\,|\,(T(X),T'(X))\bigr]}
	{P\bigl[Y=j\,|\,(T(X),T'(X))\bigr]}.
\end{equation*}
If $Q_X = Q'_X$ and $P\bigl[\mathrm{rank}\bigl(R(X)\bigr)=\ell\bigr] = 1$ is true, then it follows that
$Q[Y=y,\, X\in M] = Q'[Y=y,\, X\in M]$ for all $y \in \mathcal{Y}$ and measurable $M \subset \mathcal{X}$.
\end{theorem}
See \citet[][Theorem~4.7]{tasche2023sparse} for a proof of Theorem~\ref{th:SJS}.
The rank condition of Theorem~\ref{th:SJS} is likely to be satisfied for instance 
if $f_i(X) = \mathbf{1}_{C_i}(X)$ for
some reasonably accurate classifier $\mathbf{C}=(C_1, \ldots, C_\ell)$ as in \eqref{eq:classifier}. 
Hence identifiability of SJS ought to be given most of the time.
\vspace{-2ex}

\subsubsection{SJS and covariate shift.} As seen above, prior probability shift is
not only a special case of SJS but also implies $T$-SJS for any transformation $T$ of the features.
In contrast, examples by \citet{chen&zaharia&Zou:SJS} and \citet{tasche2023sparse} show that
covariate shift and SJS are unrelated properties in the sense that they do not
imply one another but do not exclude each other either.

For a full understanding of the relationship of covariate shift and SJS, we introduce two
further types of dataset shift. The first of these was proposed by \citet[][Definition~4.11]{tasche2023sparse}.
\begin{definition}[Conditional distribution invariance (CDI)]\label{de:cdi}
Let $T:\mathcal{X}\to \mathcal{T}$ be a measurable transformation of the feature variable $X$.
The source distribution $P$ and the target distribution $Q$ are related through 
\emph{$T$-CDI} if it holds for all $M \subset \mathcal{X}$ that
\begin{equation}\label{eq:cdi}
P[X \in M\,|\,T(X)=t] = Q[X \in M\,|\,T(X)=t]
\end{equation}
for all $t\in \mathcal{T}$ almost surely under $P_{T(X)}$ and $Q_{T(X)}$.
\end{definition}
The property CDI is interesting because in principle its presence can be evidenced by comparing 
statistics estimated from the feature observations in the training and test datasets. No label observations
are needed. Moreover, in the presence of CDI, there is basically no difference between covariate shift
and SJS, as we will see below.

The following additional type of dataset shift was introduced by \citet[][Definition~3]{chen&zaharia&Zou:SJS}.
\begin{definition}[Sparse Covariate Shift (SCS)]\label{de:SCS}
Let $T:\mathcal{X}\to \mathcal{T}$ be a measurable transformation of the feature variable $X$.
The source distribution $P$ and the target distribution $Q$ are related through 
\emph{$T$-SCS} if it holds for all $y \in \mathcal{Y}$ and $M \subset \mathcal{X}$ that
\begin{equation}\label{eq:SCS}
P[X \in M, Y=y\,|\,T(X)=t] = Q[X \in M, Y=y\,|\,T(X)=t]
\end{equation}
for all $t\in \mathcal{T}$ almost surely under $P_{T(X)}$ and $Q_{T(X)}$.
\end{definition}

The following theorem describes the interplay of SJS and covariate shift in the presence of CDI.
\begin{theorem}\label{th:cdi}
Let $T:\mathcal{X}\to \mathcal{T}$ be a measurable transformation of the feature variable $X$. 
Suppose that a source distribution $P$ and a target distribution $Q$ on $\mathcal{X}\times\mathcal{Y}$
are given. Then the following three statements hold true:
\begin{itemize}
\item[(i)] If $P$ and $Q$ are related through both $T$-CDI in the sense of
Definition~\ref{de:cdi} and covariate shift in the sense of Definition~\ref{de:covSh}, then
$P$ and $Q$ are also related through $T$-SCS in the sense of Definition~\ref{de:SCS}.
\item[(ii)] If $P$ and $Q$ are related through $T$-SCS, they are also related through
both $T$-SJS and $T$-CDI.
\item[(iii)] For given measurable functions $f_i:\mathcal{X}\to[0,\infty)$, $i = 1, \ldots, \ell$, define
the random matrix $R(X) = \bigl(R_{ij}(X)\bigr)_{i,j\in\{1, \ldots, \ell\}}$ by
\begin{equation*}
R_{ij}(X) = \frac{E_p\bigl[f_i(X)\,\mathbf{1}_{\{j\}}(Y)\,|\,T(X)\bigr]}
	{P\bigl[Y=j\,|\,T(X)\bigr]}.
\end{equation*}
Suppose that $P\bigl[\mathrm{rank}\bigl(R(X)\bigr)=\ell\bigr] = 1$ holds true. Then, if
$P$ and $Q$ are related through both $T$-SJS and $T$-CDI, they are also related
through covariate shift.
\end{itemize}
\end{theorem}
For the derivation of Theorem~\ref{th:cdi}, see Theorem~4.16 and Remark~4.18 of \citet{tasche2023sparse}.
Somewhat oversimplifying, we might summarise Theorem~\ref{th:cdi} with the following 
`equation': 
\quad $SCS\  =\ covariate\ shift\ \cap\ CDI\ =\ SJS\ \cap\ CDI.$
\vspace{-2ex}

\subsubsection{Class distribution estimation under SJS.}
\citet{chen&zaharia&Zou:SJS} proposed two methods for estimating SJS: SEES-c for the case of continuous features and
SEES-d for the case of discrete features (SEES = ``shift estimation and explanation under SJS''). 
In this paper, we briefly 
describe only an important special case of SEES-d \citep[][Eq.~(C.6)]{tasche2023sparse}
because the results presented by \citet{chen&zaharia&Zou:SJS} 
appear to suggest that SEES-d is more efficient than SEES-c. By sufficiently fine discretisation of the feature space, 
SEES-d can also be applied to continuous or mixed continuous and discrete feature settings.
\begin{subequations}
\begin{proposition}[Conditional confusion matrix approach]\label{pr:condConfusion}
Let $T:\mathcal{X}\to \mathcal{T}$ be a measurable and discrete transformation of the feature variable $X$,
i.e.\ with range $\mathcal{T} = \{t_1, \ldots, t_N\}$.
Suppose that the source distribution $P$ and a target distribution $Q$ are related through $T$-SJS in the sense
of Definition~\ref{de:SJS} and that $\mathbf{C}=(C_1, \ldots, C_\ell)$ is a classifier as in \eqref{eq:classifier}.
Then for each $t \in \mathcal{T}$, the target posterior probabilities $q_{y,t} = Q[Y=y\,|\,T(X)=t]$, 
$y \in \mathcal{Y}$, satisfy the linear equation system (with $j = 1, \ldots, \ell$)
\begin{equation}\label{eq:condConfusion}
\sum_{y=1}^\ell q_{y,t} \,P[X\in C_j\,|\,Y=y,T(X)=t] = Q[X\in C_j\,|\, T(X)=t].
\end{equation}
\end{proposition}
Once the $q_{y,t}$, $y \in \mathcal{Y}, t \in \mathcal{T},$ have been determined, by the law of total probability 
the target class prior probabilities $Q[Y=y]$ can be calculated via
\begin{equation}\label{eq:total}
Q[Y=y] = \sum_{i=1}^N q_{y,t_i}\,Q[T(X)=t_i].
\end{equation}
Therefore, Proposition~\ref{pr:condConfusion} provides a solution to the class distribution estimation 
problem under an assumption of SJS, thereby generalising the confusion matrix approach as described by
\citet[][Section~2.3.1]{saerens2002adjusting}. In particular, Proposition~\ref{pr:condConfusion} could be deployed
to check assumptions of prior probability shift. By Proposition~\ref{pr:SJS}~(i), 
prior probability shift implies $T$-SJS
for any transformation $T$. Hence, in principle, results under prior probability shift
by any suitable method of class distribution estimation must coincide with the results obtained by 
combining \eqref{eq:condConfusion} and \eqref{eq:total}, for any choice of $T$ taking discrete values.
\end{subequations}

In practice, develop the classifier on the full training dataset. Then stratify both training dataset 
and test dataset by $T$ applied to the feature (or covariate) variable $X$. 
After that, treat each of the resulting sub-samples
with the confusion matrix approach as in \citet[][Section~2.3.1]{saerens2002adjusting} to estimate 
for each $t \in \mathcal{T}$ the posterior probabilities $Q[Y=y\,|\,T(X)=t] = q_{y,t}$, $y\in\mathcal{Y}$.
Combine the posterior probabilities by means of \eqref{eq:total} to obtain estimates of the target
prior probabilities $Q[Y=y]$, $y\in\mathcal{Y}$.

Examples for possible choices of the transformation $T$ of Proposition~\ref{pr:condConfusion} might be found
in medical applications: It is plausible that the sensitivity and specificity of a test for an infection
change between training and test datasets  but that they are preserved within the strata when there is stratification 
by age group and gender. This would mean that the dataset shift can be described by $T$-sparse joint shift
with $T$ being the transformation that provides the age group and the gender of an instance (patient).

\section{Conclusions}
\label{se:Concl}

This paper provides analyses of invariance assumptions for distribution (dataset) shift, with focus
on their suitability for designing class distribution estimators. Covariate shift, factorizable joint shift,
and sparse joint shift are studied in some detail. Both the `covariate'  and the `sparse joint' types of shift
are found fit for designing class distribution estimators. In contrast, factorizable joint shift is found unsuitable 
due to lack of identifiability unless additional constraints are applied. 

Sparse joint shift (SJS) is particularly appealing for the fact that it generalises prior probability shift 
(label shift) and,
therefore, has the potential to provide meaningful estimates even in contexts where an assumption
of prior probability shift is found untenable. An open research problem is how to identify feature transformations that 
entail SJS if they cannot be identified by theoretical considerations. \citet[][Section~4.1]{chen&zaharia&Zou:SJS} 
suggested two brute-force approaches but these approaches have issues which might make their application 
questionable \citep[][Section~5]{tasche2023sparse}. 
\vspace{-2ex}

\subsubsection{Acknowledgement} The author would like to thank three anonymous reviewers for 
their useful comments and suggestions. 
\vspace{-2ex}

%
%
%
\bibliographystyle{plainnat}
{\small
\bibliography{C:/Users/Dirk/Documents/LehreForschung/Papers/Literature}
}

\addcontentsline{toc}{section}{References}

\end{document}